\documentclass[a4paper, 10pt, conference]{IEEEtran}
\IEEEoverridecommandlockouts
\usepackage{cite}
\usepackage{amsmath,amssymb,amsfonts}
\usepackage{dsfont}
\usepackage{algorithmic}
\usepackage{graphicx}
\usepackage{textcomp}
\usepackage{xcolor}
\usepackage{tikz}
\usepackage{float}
\usepackage{mathrsfs}
\usepackage{todonotes}
\usepackage{hyperref}
\usepackage{nicefrac}
\usepackage{units}
\usepackage[capitalise,nameinlink]{cleveref}
\hypersetup{colorlinks=true}
\usepackage{rotating}
\usepackage{multirow}
\usepackage{placeins}
\usepackage{caption}
\usepackage{subcaption}

\renewcommand{\Cref}[1]{\cref{#1}}
\Crefname{equation}{Eq.}{Eqs.}
\Crefname{figure}{Fig.}{Figs.}
\Crefname{tabular}{Tab.}{Tabs.}
\Crefname{table}{Tab.}{Tabs.}
\Crefname{section}{Sec.}{Secs.}

\def\BibTeX{{\rm B\kern-.05em{\sc i\kern-.025em b}\kern-.08em
    T\kern-.1667em\lower.7ex\hbox{E}\kern-.125emX}}
\begin{document}
 
\title{LU-Net: Invertible Neural Networks Based on Matrix Factorization}
\author{\IEEEauthorblockN{1\textsuperscript{st} Robin Chan\IEEEauthorrefmark{1}}
\IEEEauthorblockA{\textit{Machine Learning Group} \\
\textit{Bielefeld University}\\
Bielefeld, Germany\\
{\small \texttt{\href{mailto:rchan@techfak.uni-bielefeld.de}{rchan@techfak.uni-bielefeld.de}}}}
\and
\IEEEauthorblockN{1\textsuperscript{st} Sarina Penquitt\IEEEauthorrefmark{1}}
\IEEEauthorblockA{\textit{IZMD} \\
\textit{University of Wuppertal}\\
Wuppertal, Germany \\
{\small \texttt{\href{mailto:penquitt@uni-wuppertal.de}{penquitt@uni-wuppertal.de}}}}
\and
\IEEEauthorblockN{3\textsuperscript{rd} Hanno Gottschalk}
\IEEEauthorblockA{\textit{IZMD} \\
\textit{University of Wuppertal}\\
Wuppertal, Germany \\
{\small \texttt{\href{mailto:gottsch@uni-wuppertal.de}{hgottsch@uni-wuppertal.de}}}}
}

\newcommand{\ie}{i.e.\ }
\newcommand{\eg}{e.g.\ }
\newcommand{\cf}{cf.\ }
\newcommand{\vs}{vs.\ }
\newcommand{\wrt}{w.r.t.\ }
\newcommand{\RC}[1]{{\color{blue!50!black} #1}}
\newcommand{\HG}[1]{{\color{green!50!black} #1}}

\maketitle
\thispagestyle{plain}
\pagestyle{plain}
\begingroup\renewcommand\thefootnote{\IEEEauthorrefmark{1}}
\footnotetext{equal contribution}
\endgroup

\begin{abstract}
LU-Net is a simple and fast architecture for invertible neural networks (INN) that is based on the factorization of quadratic weight matrices $\mathsf{A=LU}$, where $\mathsf{L}$ is a lower triangular matrix with ones on the diagonal and $\mathsf{U}$ an upper triangular matrix. Instead of learning a fully occupied matrix $\mathsf{A}$, we learn $\mathsf{L}$ and $\mathsf{U}$ separately. If combined with an invertible activation function, such layers can easily be inverted whenever the diagonal entries of $\mathsf{U}$ are different from zero. Also, the computation of the determinant of the Jacobian matrix of such layers is cheap. Consequently, the LU architecture allows for cheap computation of the likelihood via the change of variables formula and can be trained according to the maximum likelihood principle. In our numerical experiments, we test the LU-net architecture as generative model on several academic datasets. We also provide a detailed comparison with conventional invertible neural networks in terms of performance, training as well as run time.  
\end{abstract}


\section{Introduction} \label{sec:intro}
In recent years, learning generative models has emerged as a major field of AI research \cite{goodfellow2014gan, kingma2014vae, dinh2017density}. In brief, given $\mathsf{N}$ i.i.d.\, examples $\mathsf{\{X_j\}_{j=1}^N}$ of a $\mathsf{\mathbb{R}^Q}$-valued random variable $\mathsf{X \sim P_X}$, the goal of a generative model is to learn the data distribution $\mathsf{P_X}$ while having only access to the sample $\mathsf{\{X_j\}_{j=1}^N}$.
Having knowledge about the underlying distribution of observed data allows for \eg identifying correlations between observations. Common downstream tasks of generative models then include 1) \emph{density estimation}, \ie what is the probability of an observation $\mathsf{x}$ under the distribution $\mathsf{P_X}$, and 2) \emph{sampling}, \ie generating novel data points $\mathsf{x_\text{new}}$ approximately following data distribution $\mathsf{P_X}$.

Early approaches include \textbf{deep Boltzmann machines} \cite{hinton2006fast, Salakhutdinov2009boltzmann}, which are energy-based models motivated from statistical mechanics and which learn unnormalized densities. The expressive power of such models is however limited and sampling from the learned distributions becomes intractable even in moderate dimensions due to the computation of a normalization constant, for which the (slow) Markov chain Monte Carlo method is often used instead.

On the contrary, \textbf{autoregressive models} \cite{larochelle2011neural,van2016pixel, papamakarios2017masked} learn tractable distributions by relying on the chain rule of probability, which allows for factorizing the density of $\mathsf{X}$ over its $\mathsf{Q}$ dimensions. In this way, the likelihood of data can be evaluated easily, yielding promising results for density estimation and sampling by training via maximum likelihood. Nonetheless, sampling using autoregressive models still remains computationally inefficient given the sequential nature of the inference step over the dimensions.

\textbf{Generative adversarial networks} (GANs) avoid the just mentioned drawback by learning a generator in a minimax game against a discriminator \cite{goodfellow2014gan, arjovsky17wasserstein}.
Here, both the generator and discriminator are represented by deep neural networks. In this way, the learning process can be conducted without an explicit formulation of the density of the distribution $\mathsf{P_X}$ and therefore can be considered as likelihood-free approach. While GANs have shown successful results on generating novel data points even in high dimensions, the absence of an expression for the density makes them unsuitable for density estimation.

Also \textbf{variational autoencoders} (VAEs) have shown fast and successful sampling results \cite{kingma2014vae, rezende2014stochastic, kingma2016improved}. These models consist of an encoder and a decoder part, which are both based on deep neural network architectures. The encoder maps data points $\mathsf{\{X_j\}_{j=1}^N}$ to lower dimensional latent representations, which define variational distributions. The decoder subsequently maps samples from these variational distributions back to input space. Both parts are trained jointly by maximizing the evidence lower bound (ELBO), \ie providing a lower bound for the density.

More recently, \textbf{diffusion models} have gained increased popularity \cite{welling2011bayesian, sohl2015deep, ho2020denoising, yang2022diffusion} given their spectacular results. In the forward pass, this type of model gradually adds Gaussian noise to data points. Then, in the backward pass the original input data is recovered from noise using deep neural networks, \ie representing the sampling direction of diffusion models.
However, sampling requires the simulation of a multi stage stochastic process, 
which is computationally slow compared to simply applying a map. Moreover, just like VAEs, diffusion models do not provide a tractable computation of the likelihood. By training using ELBO, they at least provide a lower bound for the density. 

Another type of generative models are \textbf{Normalizing flows} \cite{rezende2015variational, dinh2015nice, dinh2017density, ardizzone2018analyzing, Kingma2018glow, kobyzev2020normalizing, teshima2020coupling, grcic2021denseflow}, which allow for efficient and exact density estimation as well as sampling of high dimensional data. These models learn an invertible transformation $\mathsf{f :\mathbb{R}^Q\to\mathbb{R}^Q}$ for $\mathsf{X\sim P_X}$ such that the distribution $\mathsf{f^{-1}(Z)}$ is as close as possible to the target distribution $\mathsf{P_X}$. Here, $\mathsf{Z}$ follows a simple prior distribution $\mathsf{P_Z}$ that is easy to evaluate densities with and easy to sample from, such as \eg multivariate standard normal distribution. Hence, $\mathsf{f^{-1}}$ represents the generator for the unknown (and oftentimes complicated) distribution $\mathsf{P_X}$. As normalizing flows are naturally based on the change of variables formula 
\begin{equation}\label{eq:change-of-variables}
    \mathsf{p_X(x) = p_Z(f(x)) \cdot \big|\det\left(\mathbb{J}f(x)\right)\big|},
\end{equation}
with $\mathsf{X=f^{-1}(Z)}$, this expression can also be used for exact probability density evaluation and likelihood based learning.

In this way, normalizing flows offer impressive generative performance along with an explicit and tractable expression of the density. Despite the coupling layers in normalizing flows being rather specific, it has been shown recently that their expressive power is universal for target measures having a density on the target space $\mathsf{\mathbb{R}^D}$ \cite{teshima2020coupling}. Note however that multiple such coupling layers are usually chained for an expressive generative model. These layers are typically parameterized by the outputs of neural networks and therefore normalizing flows can require considerable computational resources.

In this work, we propose \textbf{LU-Net}: an alternative to existing invertible neural network (INN) architectures motivated by the positive properties of normalizing flows. The major advantage of LU-Net is the simplicity of its design. It is based on the elementary insight that a fully connected layer of a feed forward neural network is a bijective map from $\mathsf{\mathbb{R}^Q\to\mathbb{R}^D}$, if and only if (a) the weight matrix is quadratic, \ie $\mathsf{Q=D}$, (b) the weight matrix is of full rank and (c) the employed activation function maps $\mathbb{R}$ to itself bijectively. A straight forward idea is then to fully compose a neural network of such invertible layers and learn a transformation $\mathsf{f}$ as in \Cref{eq:change-of-variables}.

However, it can be computationally expensive to actually invert the just described INN, especially if required during training. Due to the size of weight matrices and the fact that they can be fully occupied, the computational complexity for the inversion scales as $\mathcal{O}\mathsf{(M\times D^3)}$, where $\mathsf{M}$ is the depth and $\mathsf{D}$ the width of the proposed INN. The same complexity also holds for the computation of the determinant, which occurs in the computation of the likelihood. Both tasks, inversion and computation of the determinant, can be based on the LU-decomposition, where a quadratic matrix is factorized in a lower triangular matrix with ones on the diagonal and an upper triangular matrix with nonzero diagonal entries. Using this factorization inversion becomes of order $\mathcal{O}\mathsf{(M\times D^2)}$ and computation of the determinant of $\mathcal{O}\mathsf{(M\times D)}$. In the LU-Net architecture, we therefore enforce the weight matrices to be of lower or upper triangular shape and keep this shape fixed during the entire training process. Consequently, a single fully connected layer $\mathsf{x\mapsto Ax+b}$ is decomposed into two layers $\mathsf{x\mapsto Ux\mapsto LUx+b}$, where the weight matrices $\mathsf{L}$ and $\mathsf{U}$ are ``masked'' on the  upper and lower triangular positions, respectively.     

This simple architecture of LU-Net however comes with a limitation, that is universal approximation. 
The classical type of universal approximation theorems deal with a fixed depth and an arbitrary width of neural networks \cite{Cybenko1989,Hornik1991, Leshno1993nonpoly}. Obviously, these theorems cannot be applied to LU-Net due its bijectivity constraint. More recent universal approximation theorems deal with a fixed width and an arbitrary depth \cite{yarotsky2017error, zhou2017, kidger2020universal, park2021minimum}. But even in the weak sense of $\mathsf{L^p}$-distances, a network requires a width of at least $\mathsf{D+1}$ \cite{park2021minimum} to be a universal approximator. Thus, the LU-Net just misses the property by one dimension. 

One could imagine that the missing dimension in the width of LU-Net becomes less relevant the higher dimensional the problem is, so that the difference between width $\mathsf{D}$ and $\mathsf{D+1}$ becomes marginal. Nevertheless, also for dimensions as low as $\mathsf{D=2}$, we provide numerical evidence that the expressivity of LU-Net still achieves reasonable quality in density estimation.

Moreover, we also present reasonable results using LU-Net for the popular task of generative modeling of images, which includes density estimation and sampling. In a quantitative comparison as suggested by \cite{Theis2016a}, LU-Net achieves a consistent advantage in terms of the negative log likelihood metric when compared to the widely used RealNVP INN architecture \cite{dinh2017density} with about the same number of parameters. In our experiments we further observe that training LU-Net is computationally considerably cheaper than training the just mentioned coupling layer based normalizing flow.
This also points to the particular suitability of LU-Net as base model for rapid prototyping.

Overall, LU-Net provides a simple and efficient framework of an INN. Due to its simplicity, this model is applicable to a variety of problems with data of different forms. This is in contrast to other generative models, \eg normalizing flows, which are often particularly designed for a specific application such as image generation.

The content of this paper is structured as follows: in \Cref{sec:LU-Net} we describe the LU-Net architecture with details on computing the density and likelihood. Numerical results including experiments on the image datasets MNIST \cite{lecun2010mnist} and Fashion-MNIST \cite{xiao2017fashion} follow in \Cref{sec:experiments}. Finally, in \Cref{sec:conclusion} we conclude our article and give recommendations for future research directions. 

Additionally, we report evaluations on condition numbers of the LU-layers and on the closeness of the distribution in the normalizing direction of LU-Net to a multivariate standard normal distribution in \Cref{sec:condition} and \Cref{sec:normality}, respectively. 
The entire code for the numerical experiments presented in this paper is publicly available on GitHub: \href{https://github.com/spenquitt/LU-Net-Invertible-Neural-Networks}{https://github.com/spenquitt/LU-Net-Invertible-Neural-Networks}.

\section{LU-Net Architecture and the Likelihood}
\label{sec:LU-Net}

Fully connected neural networks are the most basic models for deep learning, which can be applied to various applications. This generality property is what we also aim for invertible neural networks in the context of probabilistic generative modeling. To this end, we have to ensure that the model is bijective and that the inversion as well as computation of the likelihood are both tractable.

The bijectivity constraint in fully connected neural networks can easily be fulfilled by using a bijective activation function and restricting the weight matrices to be quadratic and of full rank. However, the inversion and computation of the determinant for the computation of the likelihood, \cf \Cref{eq:change-of-variables}, of fully occupied matrices remain computationally expensive with a cubic complexity. To address this problem, we propose to directly learn the LU factorization replacing the weight matrices in fully connected layers without a loss in model capacity. This forms the building block of our proposed LU-Net, which we explain in more details in what follows.

\subsection{LU Factorization}
    The LU factorization is a common method to decompose any square matrix $\mathsf{A} \in \mathsf{\mathbb{R}^{D \times D}}$ into a lower triangular matrix $\mathsf{L} \in \mathsf{\mathbb{R}^{D \times D}}$ with ones on the diagonal and in an upper triangular matrix $\mathsf{U} \in \mathsf{\mathbb{R}^{D \times D}}$ with non-zero diagonal entries. Then, the matrix $\mathsf{A}$ can be rewritten as
    
    \begin{tikzpicture}
      \node [anchor=center, text width=.9\linewidth ] at (0,0) {
        \begin{minipage}{\textwidth}
            \begin{align} \label{eq:lu-factorization}
                \mathsf{A=LU} =
                \left(
                \begin{array}{ccc}
                    1   & &  \\
                        & \ddots & \\
                        & & 1
                \end{array}
                \right) 
                \left(
                \begin{array}{ccc}
                      \mathsf{u_{1,1}} & & \\
                        & \ddots  & \\
                       & & \!\!\!\mathsf{u_{D,D}}
                \end{array}
                \right)
            \end{align}
        \end{minipage}};
        \draw (-1.75,-.75) -- (-.75,-.75) -- (-1.75,0) -- (-1.75,-.75) ;
        \draw (2.75,.5) -- (1.75,.5) -- (2.75,-.25) -- (2.75,.5) ;
        \node [anchor=center] at (-.35,.35) {\text{\huge 0}} ;
        \node [anchor=center] at (1.,-.55) {\text{\huge 0}} ;
    \end{tikzpicture}\\
    with $\mathsf{u_{i,i}} \neq 0 ~ \forall ~ \mathsf{i=1,\ldots,D}$.
    One popular application of this factorization is \eg solving linear equation systems. We use this factorization to replace fully occupied weight matrices in fully connected layers of neural networks. Given \Cref{eq:lu-factorization} the computation of $\mathsf{A}^{-1}$ and $\det(\mathsf{A})$ become notably more efficient with time complexities of order $\mathcal{O}\mathsf{(D^2)}$ and $\mathcal{O}\mathsf{(D)}$, respectively, instead of $\mathcal{O}\mathsf{(D^3)}$ without the LU decomposition.

\subsection{LU-Net Architecture}

In the following, let $\mathsf{x\in\mathbb{R}^D}$ denote some $\mathsf{D}$-dimensional input, \ie each layer of LU-Net will be a map $\mathsf{\mathbb{R}^D} \to \mathsf{\mathbb{R}^D}$. Further, let $\mathsf{M \geq 2}$ specify the number of LU layers.

\begin{figure}
    \centering
    \captionsetup[subfloat]{margin=10pt,format=hang}
    \begin{subfigure}{0.48\linewidth}
        \resizebox{\linewidth}{!}{\tikzstyle{neuron}=[thick,draw=black,fill=purple!50,circle,minimum size=2]

\begin{tikzpicture}[x=2.2cm,y=1.4cm]
    \foreach \N [count=\lay,remember={\N as \Nprev (initially 0);}] in {7,7,7,7,7}{ 
        \foreach \i [evaluate={\y=\N/2-\i/3.5; \x=\lay/2; \prev=int(\lay-1);}] in {1,...,\N}{ 
            \node[neuron] (N\lay-\i) at (\x,\y) {};
            \ifnum\Nprev > 0
                \ifnum\lay = 2 
                    \foreach \j in {1,...,\i}{ 
                        \draw[thick] (N\prev-\j) -- (N\lay-\i);
                    }
                \fi
                \ifnum\lay = 3 
                    \foreach \j in {\i,...,\Nprev}{ 
                        \draw[thick] (N\prev-\j) -- (N\lay-\i);
                    }
                \fi
                \ifnum\lay > 3 
                    \draw[thick] (N\prev-\i) -- (N\lay-\i);
                \fi
            \fi
        }
    }
    \draw[thick, -latex] (.5,1.25) -- (2.5,1.25);
    \node[anchor=center] at (0.75, 3.5) {$\mathsf{U}$};
    \node[anchor=center] at (1.25, 3.5) {$\mathsf{L}$};
    \node[anchor=center] at (1.75, 3.5) {$\mathsf{b}$};
    \node[anchor=center] at (2.25, 3.5) {$\mathsf{\Phi}$};
\end{tikzpicture}}
        \caption{normalizing sequence: \\ 
            $\mathsf{x \mapsto Ux}$ \\
            $\mathsf{\mapsto LUx}$ \\ 
            $\mathsf{\mapsto LUx+b}$ \\ 
            $\mathsf{\mapsto \Phi(LUx+b)}$ \\
            $=\mathsf{f(x) =: z}$
        }
        \label{fig:normalizing_direction}
    \end{subfigure}\hfill%
    \begin{subfigure}{0.48\linewidth}
        \resizebox{\linewidth}{!}{\tikzstyle{neuron}=[thick,draw=black,fill=purple!50,circle,minimum size=2]

\begin{tikzpicture}[x=2.2cm,y=1.4cm]
    \foreach \N [count=\lay,remember={\N as \Nprev (initially 0);}] in {7,7,7,7,7}{ 
        \foreach \i [evaluate={\y=\N/2-\i/3.5; \x=\lay/2; \prev=int(\lay-1);}] in {1,...,\N}{ 
            \node[neuron] (N\lay-\i) at (\x,\y) {};
            
            \ifnum\Nprev > 0
                 \ifnum\lay < 4 
                    \draw[thick] (N\prev-\i) -- (N\lay-\i);
                \fi
                \ifnum\lay = 4 
                    \foreach \j in {\i,...,\Nprev}{ 
                        \draw[thick] (N\prev-\j) -- (N\lay-\i);
                    }
                \fi
                \ifnum\lay = 5 
                    \foreach \j in {1,...,\i}{ 
                        \draw[thick] (N\prev-\j) -- (N\lay-\i);
                    }
                \fi
            \fi
        }
    }
    \draw[thick, -latex] (.5,1.25) -- (2.5,1.25);
    \node[anchor=center] at (0.75, 3.5) {$\mathsf{\Phi^{-1}}$};
    \node[anchor=center] at (1.25, 3.5) {$\mathsf{-b}$};
    \node[anchor=center] at (1.75, 3.5) {$\mathsf{L^{-1}}$};
    \node[anchor=center] at (2.25, 3.5) {$\mathsf{U^{-1}}$};
\end{tikzpicture}}
        \caption{generating sequence: \\ 
            $\mathsf{z \mapsto \Phi^{-1}(z)}$ \\
            $\mathsf{\mapsto \Phi^{-1}(z)-b}$ \\ 
            $\mathsf{\mapsto L^{-1}(\Phi^{-1}(z)-b)}$ \\ 
            $\mathsf{\mapsto U^{-1}L^{-1}(\Phi^{-1}(z)-b)}$ \\
            $=\mathsf{f^{-1}(z)=x}$
        }
        \label{fig:generating_direction}
    \end{subfigure}
    \caption{Illustration of the LU-Net design with one $\mathsf{LU}$ layer. Here, (a) represents the forward or ``normalizing'' direction and (b) the reversed or ``generating'' direction. Note that the weights are shared and that $\mathsf{\left(f^{-1} \circ f \right)(x) = x}$.}
    \label{fig:lu-layer-design}
\end{figure}
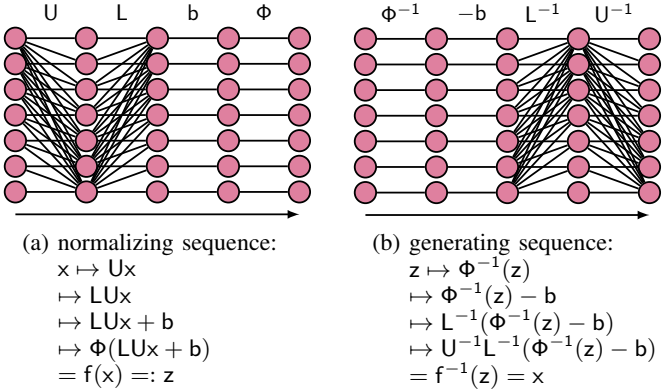

In each layer we apply two linear transformations to some input $\mathsf{x}$, which is then followed by a (non-linear) transformation. More precisely, in one LU-Net layer we apply the sequence $\mathsf{x\mapsto Ux\mapsto LUx+b \mapsto \Phi(LUx+b)}$, yielding
\begin{align*}\
    \mathsf{f^{(m)}(x)}
    &=\mathsf{\Phi^{(m)}\left(L^{(m)}U^{(m)}x+b^{(m)}\right)}\\
    &=\left(
        \begin{array}{c}
        \mathsf{\phi^{(m)}\left((L^{(m)}U^{(m)}x)_1+b_1^{(m)}\right)}\\
        \vdots\\
        \mathsf{\phi^{(m)}\left((L^{(m)}U^{(m)}x)_D+b_D^{(m)}\right)}
        \end{array}
    \right)
\end{align*}
as the output of the $\mathsf{m}$-th LU layer for all $\mathsf{m=1,\ldots,M}$, see also \Cref{fig:normalizing_direction}. Here, $\mathsf{L}$ and $\mathsf{U}$ contain learnable weights and have a fixed shape as in \Cref{eq:lu-factorization}. $\mathsf{b \in \mathbb{R}^D}$ is the learnable bias and 
\begin{equation}\label{eq:activations_layer}
    \mathsf{\Phi\left(x\right)= \left( \phi(x_1), \ldots, \phi(x_D) \right)^\intercal },~ \mathsf{\phi:\mathbb{R} \to \mathbb{R}}
\end{equation}
the real-valued activation function that is applied element-wise.

Chaining multiple such LU layers together, we obtain
\begin{equation*}
    \mathsf{f(x)=\left(f^{(M)}\circ f^{(M-1)}\circ\cdots\circ f^{(1)}\right)(x)},~ \mathsf{f :\mathbb{R}^D \to \mathbb{R}^D}
\end{equation*}
as overall expression for the forward direction of LU-Net, \ie the ``normalizing`' direction, \cf \Cref{eq:change-of-variables}.

\subsection{Activation Functions}

\begin{figure}
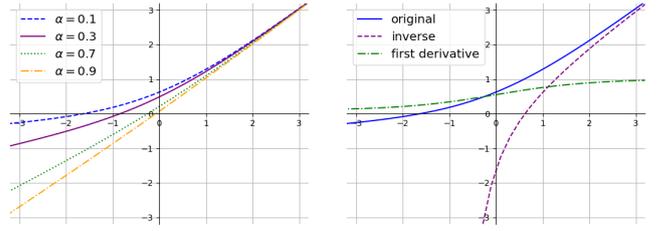

\centering
\captionsetup[subfloat]{margin=10pt,format=hang}
\begin{subfigure}{0.5\linewidth}
    \includegraphics[width=\textwidth, trim=1cm 0 1cm 0, clip]{images/Activation/leaky_plot_slope.pdf}
    \caption{leaky softplus function\\ with varying slopes $\mathsf{\alpha}$}
    \label{fig:LeakySlope}
\end{subfigure}\hfill%
\begin{subfigure}{0.5\linewidth}
    \includegraphics[width=\textwidth, trim=1cm 0 1cm 0, clip]{images/Activation/leaky_plot.pdf}
    \caption{leaky softplus ($\mathsf{\alpha=0.1}$), its derivative and inverse}
    \label{fig:Leaky}
\end{subfigure}
\caption{Illustration of (a) different variants of the leaky softplus function as well as (b) its derivative and inverse. We refer to \Cref{eq:leaky_softplus} for the formula of the leaky softplus function.} 
\label{fig:leaky-softplus}
\end{figure}
 
The choice of activation function $\mathsf{\phi}$ is crucial for the invertibility but also flexibility of our model $\mathsf{f}$.
In LU-Net we employ a ReLU-like non-linear activation function, which we term \emph{leaky softplus} and define as
\begin{equation}\label{eq:leaky_softplus}
    \mathsf{\text{LeakySoftplus}(x) := \alpha x + (1-\alpha)\ln(1+\exp(x))}
\end{equation}
with $\mathsf{\alpha\in (0,1)}$, see also \Cref{fig:leaky-softplus}.
As the name suggests, this is a combination of the leaky ReLU and softplus function. Although these two latter functions are both invertible, they have practical drawbacks. Equipped with leaky ReLU the LU-Net would suffer from a vanishing gradient in the backpropgation step during training (as the second derivative will be needed when maximizing the likelihood, \cf \Cref{eq:change-of-variables}). This is not the case for the softplus function, however the domain of its inverse only contains $\mathbb{R}_+$, the positive real values, thus clearly restricting the possible outputs of LU layers. 

The leaky softplus function circumvents those aforementioned limitations, which is why we choose this activation function for all hidden layers in our proposed LU-Net, \ie $\mathsf{\phi^{(m)}(x) = \text{LeakySoftplus}(x)}$ with $\mathsf{\alpha=0.1}$ for all $\mathsf{m=1,\ldots,M-1}$. As we will deal with regression problems, we employ no activation in the final layer, \ie $\mathsf{\phi^{(M)}(x) = x}$ is the identity map.

\subsection{Inverse LU-Net}\label{sec:inverse-lu-net}

If all previously described requirements for the inversion of LU-Net are fulfilled, each layer can be reversed as
	\begin{equation*}
		\mathsf{{f^{(m)}}^{-1}(z)={U^{(m)}}^{-1}{L^{(m)}}^{-1}\left(\Phi^{-1}(z)-b^{(m)}\right)}
	\end{equation*}
for some input $\mathsf{z\in\mathbb{R}^D}$ and for all LU layers $\mathsf{m=1,\dotsc,M}$. 

Then, the overall expression for the reversed LU-Net $\mathsf{f^{-1}:\mathbb{R}^D\to\mathbb{R}^D}$ is given by
	\begin{equation*}
		\mathsf{f^{-1}(z)=\left({f^{(1)}}^{-1}\circ\cdots\circ {f^{(M-1)}}^{-1}\circ {f^{(M)}}^{-1}\right)(z)}
	\end{equation*}
and represents the ``generating direction'', see also \Cref{fig:generating_direction}. 

Note that both $\mathsf{f}$ and $\mathsf{f^{-1}}$ share their weights and it holds $\mathsf{\left(f^{-1} \circ f \right)(x) = x}$ for any input $\mathsf{x \in \mathbb{R}^D}$ even without any training.

\subsection{Training via Maximum Likelihood}
Given a dataset $\mathsf{\mathcal{D}=\{x^{(n)}\}_{n=1}^N}$, containing $\mathsf{N}$ independently drawn examples of some random variable $\mathsf{X \sim P_X}$, our training objective is then to maximize the likelihood
	
	\begin{equation} \label{eq:likelihood_x}
		\mathsf{\mathscr{L}(\theta|\mathcal{D})=\prod_{n=1}^{N}p_X(x^{(n)})}
	\end{equation}
where $\mathsf{p_X:\mathbb{R}^D\to\mathbb{R}}$ denotes the (unknown) probability density function corresponding to the target distribution $\mathsf{P_X}$ and $\mathsf{\theta= \{ U^{(m)}, L^{(m)}, b^{(m)} \}_{m=1}^M}$ the set of model parameters of LU-Net.
By defining the model function $\mathsf{f(\cdot|\theta)}$ of LU-Net to be the invertible transform such that $\mathsf{X=f^{-1}(Z|\theta) \Leftrightarrow Z = f(X|\theta)}$, where $\mathsf{Z\sim P_Z}$ is another random variable following a simple prior distribution, we can use the change of variables formula to rewrite the expression in \Cref{eq:likelihood_x} to
	\begin{equation}
    \label{eq:likelihood}
		\mathsf{\mathscr{L}(\theta|\mathcal{D})=\prod_{n=1}^{N}p_Z\left(f(x^{(n)}|\theta)\right) \big|\det\left(\mathbb{J}f(x^{(n)}|\theta)\right)\big|} ~.
	\end{equation}	
	
As in normalizing flows, we choose $\mathsf{P_Z}$ to be a $\mathsf{D}$-multivariate standard normal distribution with probability density function 
\begin{equation}\label{eq:pdf-normal}
    \mathsf{p(z) = \frac{1}{\sqrt{(2\pi)^D}}\, \exp\left(-\frac{1}{2} z^\intercal z\right)}, ~ \mathsf{z\in\mathbb{R}^D}.
\end{equation}
    
Further, given the fact that the determinant of a triangular matrix is the product of its diagonal entries and given the chain rule of calculus, for each LU layer $\mathsf{m=1,\ldots,M}$ of LU-Net it applies
\begin{align}\label{eq:determinant}
    \begin{split}
    & \mathsf{\big|\det\left(\mathbb{J}f^{(m)}(x)\right)\big|} \\
    = & \mathsf{\big| \prod_{m=1}^M \prod_{d=1}^D \phi^{\prime(m)}\left((L^{(m)}U^{(m)}x)_d+b_d^{(m)}\right) \cdot u_{d,d}^{(m)} \big| }
    \end{split}
\end{align}
with $\mathsf{\phi^{\prime(m)}}$ being the derivative of the $\mathsf{m}$-th LU layer's activation function $\mathsf{\phi^{(m)}}$.

Considering now again the chain rule of calculus and taking into account \Cref{eq:likelihood}, \Cref{eq:pdf-normal} as well as \Cref{eq:determinant}, we obtain the following final expression for the negative log likelihood as training loss function:
\begin{align}\label{eq:log-likelihood}
\begin{split}
    & \mathsf{-\ln\mathscr{L}(\theta|\mathcal{D})}\\
    =& \phantom{-} \mathsf{\frac{1}{2}\cdot N\cdot D\cdot\ln(2\pi)+\frac{1}{2}\sum_{n=1}^{N}\sum_{d=1}^D f_d(x^{(n)}|\theta)^2 } \\
    & \mathsf{-\sum_{n=1}^{N}\sum_{m=1}^M\sum_{d=1}^D \ln \phi^{\prime(m)}\left((L^{(m)}U^{(m)}x^{(n)})_d+b_d^{(m)}\right)} \\
    & \mathsf{-N \cdot \sum_{m=1}^M\sum_{d=1}^D \ln \big| u_{d,d}^{(m)} \big|} \rightarrow \min ~.
\end{split}
\end{align}
Note that for each hidden LU layer $\mathsf{m=1,\ldots,M-1}$ in LU-Net
\begin{equation*}
    \mathsf{\phi^{\prime(m)}(x)=\alpha +\frac{1-\alpha}{1+\exp(-x)} = \alpha + (1-\alpha)\, \sigma(x)}
\end{equation*}
with $\mathsf{\sigma: \mathbb{R} \to \mathbb{R}}$ being the logistic sigmoid function. For the final layer the derivative is constant with $ \mathsf{\phi^{\prime(M)}=1}$.

\section{LU-Net Experiments}\label{sec:experiments}
In this section we present extensive experiments with LU-Net, which were conducted in different settings. As a toy example, we apply LU-Net to learn a two-dimensional Gaussian mixture. Next, we apply LU-Net to the image datasets MNIST \cite{lecun2010mnist} as well as the more challenging Fashion-MNIST \cite{xiao2017fashion}. We evaluate LU-Net as density estimator and also have look at the sampling quality as generator.

\subsection{Training goal and evaluation}

In general, our goal is to learn a target distribution $\mathsf{P_X}$ given only samples which are produced by its data generating process. To this end, we attempt to train our model distribution given by LU-Net to be as close as possible to the data distribution, \ie the empirical distribution provided by examples $\mathsf{\mathcal{D}=\{x^{(n)}\}_{n=1}^N}$ of the random variable $\mathsf{X \sim P_X}$.

One common way then to quantify the closeness between two distributions $\mathsf{P}$ and $\mathsf{Q}$ with probability density functions $\mathsf{p}$ and $\mathsf{q}$, respectively, is via the Kullback-Leibler divergence
\begin{equation*}\label{eq:kl-divergence}
    \mathsf{D_{KL}(P\vert\vert Q) = \mathbb{E}_{X \sim P} \left[ \ln \frac{p(X)}{q(X)} \right] = \int_\mathbb{R} p(x) \ln \frac{p(x)}{q(x)}dx }.
\end{equation*}
It is well known that maximizing the likelihood on the dataset $\mathcal{D}$, as presented in \Cref{eq:log-likelihood}, asymptotically amounts to minimizing the Kullback-Leibler divergence between the target distribution and model distribution \cite{arjovsky17wasserstein, chan2022detecting}, which in our case are defined by $\mathsf{P_X}$ and LU-Net, respectively.
For this reason, the negative log likelihood (NLL) is not only used as loss function for training, but also as the standard metric to measure the density estimation capabilities of probabilistic generative models \cite{Theis2016a}.

\begin{table}
\setlength{\tabcolsep}{5pt}
\begin{tabular}{|c|c|c|c|}
 \hline
 \multicolumn{4}{|c|}{Gaussian mixture negative log likelihood} \\
 \hline
  \hline
 \text{\# LU Layers} & 2 & 3 & 5 \\
 \hline
 Test NLL $\downarrow$ & 3.4024 $\pm$ 0.3403 & 2.6765 $\pm$ 0.3607 & 1.8665 $\pm$ 0.2875\\
\hline
\hline
\end{tabular}

\begin{tabular}{|c|c|c|}
 \hline
 \text{\# LU Layers} & 8 & 12\\
 \hline
 Test NLL $\downarrow$ & 1.4633 $\pm$ 0.2132 & 1.0848 $\pm$ 0.3239\\
\hline
\end{tabular}
\caption{Test results of LU-Net on the two-dimensional Gaussian mixture toy problem.} 
\label{tab:GaussionMixture}
\end{table}

\begin{figure}[h]
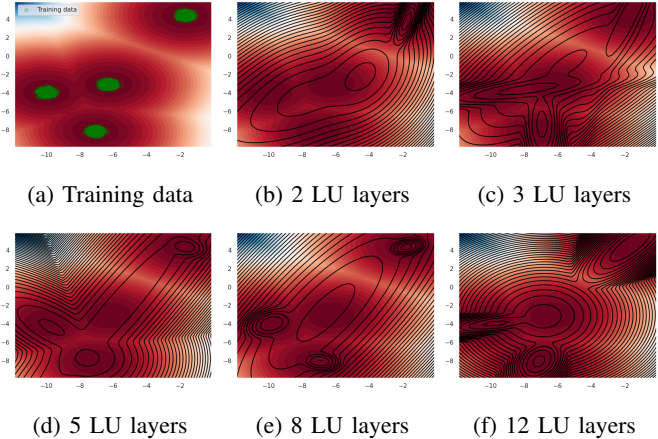

\centering
\begin{subfigure}{0.33\linewidth}
    \includegraphics[width=\textwidth, trim=1cm 0 1cm 0, clip]{images/Gaussian/Generation/Contour/contour_train.pdf}
    \caption{Training data}
    \label{fig:GenerationTrain}
\end{subfigure}%
\begin{subfigure}{0.33\linewidth}
    \includegraphics[width=\textwidth, trim=1cm 0 1cm 0, clip]{images/Gaussian/Generation/Contour/contour_2_31.pdf}
    \caption{2 LU layers}
    \label{fig:Generation2Layers}
\end{subfigure}%
\begin{subfigure}{0.33\linewidth}
    \includegraphics[width=\textwidth, trim=1cm 0 1cm 0, clip]{images/Gaussian/Generation/Contour/contour_3_52.pdf}
    \caption{3 LU layers}
    \label{fig:Generation3Layers}
\end{subfigure}\\
\begin{subfigure}{0.33\linewidth}
    \includegraphics[width=\textwidth, trim=1cm 0 1cm 0, clip]{images/Gaussian/Generation/Contour/contour_5_56.pdf}
    \caption{5 LU layers}
    \label{fig:Generation5Layers}
\end{subfigure}%
\begin{subfigure}{0.33\linewidth}
    \includegraphics[width=\textwidth, trim=1cm 0 1cm 0, clip]{images/Gaussian/Generation/Contour/contour_8_33.pdf}
    \caption{8 LU layers}
    \label{fig:Generation8Layers}
\end{subfigure}%
\begin{subfigure}{0.33\linewidth}
    \includegraphics[width=\textwidth, trim=1cm 0 1cm 0, clip]{images/Gaussian/Generation/Contour/contour_12_79.pdf}
    \caption{12 LU layers}
    \label{fig:Generation12Layers}
\end{subfigure}%
\caption{The LU-Net with different numbers of LU layers applied to a two-dimensional Gaussian mixture. The heatmaps indicate the ground truth densities and the level curves the learned densities given by training LU-Net.} 
\label{fig:figGaussianGeneration}
\end{figure}

\subsection{Gaussian mixture}
\subsubsection{Experimental setup}

To begin, we create a dataset consisting of 10,000 sampled two-dimensional Gaussian data points at four different centers with a standard deviation of $0.2$ of which we use 9,000 for training LU-Net in different configurations, see \Cref{fig:GenerationTrain}. More precisely, we train five LU-Nets in total comprising 2, 3, 5, 8, and 12 hidden LU layers with a final LU output layer eac for 10, 20, 30, 35, and 40 epochs, respectively. As optimization algorithm we use stochastic gradient descent with a momentum term of 0.9. We start with a learning rate of $1.0$ that decays by 0.9 in each training epoch. Further, we clip the gradient to a maximal length of 1 \wrt the absolute-value norm, which empirically have shown to stabilize the training process.

\subsubsection{Results}

In \Cref{tab:GaussionMixture} we report the negative log likelihood of LU-Net on the 1,000 holdout test data points. In \Cref{fig:Generation2Layers} -- \Cref{fig:Generation12Layers} we provide visualizations of the learned and ground truth density. Generally, we observe that by stacking more LU layers the model function becomes more flexible, which is in line with \cite{park2021minimum} stating that deeper neural network have increased capacity. In our toy experiments the LU-Net with 12 layers achieves the best result with an averaged NLL of 1.0848. This is visible in the visualization in \Cref{fig:Generation12Layers}, clearly showing having learned modes in proximity of the true centers of the target Gaussian mixture. We conclude that in practice depth can increase the expressive power of LU-Net. 

\begin{table}[h]
\centering
\begin{tabular}{|c|c|c|c|}
    \hline
    \multicolumn{2}{|c|}{MNIST Test NLL} & \multicolumn{2}{|c|}{Fashion-MNIST Test NLL} \\
    \hline\hline
    Class & Bits / Pixel $\downarrow$ & Class & Bits / Pixel $\downarrow$ \\
    \hline
    Number 0 & 2.7180 $\pm$ 0.0284 & T-Shirt    & 3.7726 $\pm$ 3.1228 \\
    Number 1 & 2.4795 $\pm$ 0.0125 & Trousers   & 7.2577 $\pm$ 4.9085 \\
    Number 2 & 2.9395 $\pm$ 0.0997 & Pullover   & 2.4018 $\pm$ 0.1091\\
    Number 3 & 2.7465 $\pm$ 0.0062 & Dress      & 2.5337 $\pm$ 0.0127 \\
    Number 4 & 2.7760 $\pm$ 0.0114 & Coat       & 3.3492 $\pm$ 1.6847 \\
    Number 5 & 2.7489 $\pm$ 0.0175 & Sandal     & 2.8861 $\pm$ 0.0173 \\
    Number 6 & 2.8591 $\pm$ 0.8279 & Shirt      & 2.7209 $\pm$ 1.1560 \\
    Number 7 & 2.7930 $\pm$ 0.0511 & Sneaker    & 3.6077 $\pm$ 0.0071 \\
    Number 8 & 2.7916 $\pm$ 0.0081 & Bag        & 4.3983 $\pm$ 2.0313\\
    Number 9 & 2.6576 $\pm$ 0.0187 & Ankle Boot & 4.4405 $\pm$ 0.1762 \\
    \hline\hline
    Average & 2.7480 $\pm$ 0.2931 & Average & 3.7368 $\pm$ 2.4632\\
    \hline 
\end{tabular}
\caption{Class-wise negative log likelihood (NLL) when applying LU-Net to MNIST and Fashion-MNIST test dataset. The results are averaged over 30 runs. Note that the NLL is reported in bits per pixel (sometimes also called bits per dimension or simply bpd), which is the NLL with logarithm base 2 averaged over all pixels.} 
\label{tab:nll-mnist}
\end{table}

\subsection{MNIST and Fashion-MNIST} \label{sec:exp_mnist}
\subsubsection{Data preprocessing}
The two publicly available datasets MNIST \cite{lecun2010mnist} and Fashion-MNIST \cite{xiao2017fashion} consist of gray-scaled images of resolution $28\times 28$ pixels. These images are stored in 8-bit integers, \ie each pixel can take on a brightness value from $\{ 0,1\ldots,255 \}$. Modeling such a discrete data distribution with a continuous model distribution (as we do with LU-Net by choosing a Gaussian prior, \cf \Cref{eq:pdf-normal}), could lead to arbitrarily high likelihood values, since arbitrarily narrow and high densities could be placed as spikes on each of the discrete brightness values. This practice would make the evaluation via the NLL not comparable and thus meaningless. 

Therefore, it is best practice in generative modeling to add real-valued uniform noise $\mathsf{u \sim U(0,1)}$, $ \mathsf{u \in [0,1)}$ to each pixel of the images in order to dequantize the discrete data \cite{Uria2013rnade, Theis2016a}. It turns out that the likelihood of the continuous model on the dequantized data is upper bounded by the likelihood on the original image data \cite{Theis2016a, Ho2019flow++}. Consequently, maximizing the likelihood on $\mathsf{x+u}$ will also maximize the likelihood on the original input $\mathsf{x}$. This also makes the NLL on the dequantized data a comparable performance measure with $\text{NLL}>0$ for any probabilistic generative model dealing with images. Note that the just described non-deterministic preprocessing step can easily be reverted by simply rounding off.

As additional preprocessing steps we normalize the dequantized pixel values to the unit interval $[0,1]$ by dividing by 256, and apply the logit function to transform the data distribution to a Gaussian-like shape. This output then represents the input to LU-Net. Again, these  preprocessing steps can easily be reversed by applying the inverse of logit, \ie the logistic sigmoid function, and by multiplying by 256, respectively. 

\begin{figure}
    \centering
    \begin{subfigure}{.24\textwidth}
    \resizebox{\linewidth}{!}{
    \begin{tikzpicture}
    \node[inner sep=0pt] at (0,0) {\includegraphics{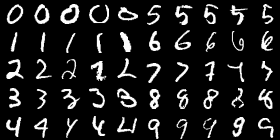}};
    \draw[white] (.05,-2.5) -- (.05,2.5);
    \end{tikzpicture}}
    \caption{MNIST}
    \end{subfigure}%
    \begin{subfigure}{.24\textwidth}
    \resizebox{\linewidth}{!}{
    \begin{tikzpicture}
    \node[inner sep=0pt] at (0,0) {\includegraphics{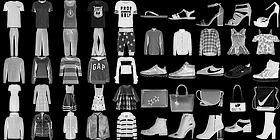}};
    \draw[white] (.0,-2.5) -- (.0,2.5);
    \end{tikzpicture}}
    \caption{Fashion MNIST}    \label{fig:density-fashionmnist}
    \end{subfigure}
    \caption{Unseen original test samples of (a) MNIST and (b) Fashion MNIST. These images are ordered class-wise and in decreasing likelihood from left to right as estimated by LU-Net. See also \Cref{tab:nll-mnist} for the class names.}
    \label{fig:density_mnist_fashionmnist}
\end{figure}

\subsubsection{Experimental setup}

We conduct the experiments on MNIST and Fashion-MNIST with an LU-Net consisting of three hidden LU layers and a final LU output layer. The model is trained for a maximum of 40 epochs and conditioned on each class of MNIST and Fashion-MNIST. As optimization algorithm, we stick to gradient descent with a momentum parameter of 0.9. We start with a learning rate of 0.6 that decays by 0.5 every three epochs. Further, we clip the gradients to a maximal length of 1 \wrt Euclidean norm, which empirically have shown to stabilize the training process as well as regularizing the weights. As extra loss weighting, we add a factor of $\mathsf{\gamma=100}$ to the sum of the diagonal entries in the NLL loss function, see \Cref{eq:log-likelihood}, which has considerably improved the convergence speed of the training process.

\subsubsection{Results}
\begin{figure}
    \centering
    \begin{subfigure}{.24\textwidth}
    \resizebox{\linewidth}{!}{
    \begin{tikzpicture}
    \node[inner sep=0pt] at (0,0) {\includegraphics{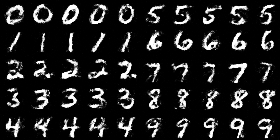}};
    \draw[white] (.05,-2.5) -- (.05,2.5);
    \end{tikzpicture}}
    \caption{MNIST}\label{fig:sampling_mnist}
    \end{subfigure}%
    \begin{subfigure}{.24\textwidth}
    \resizebox{\linewidth}{!}{
    \begin{tikzpicture}
    \node[inner sep=0pt] at (0,0) {\includegraphics{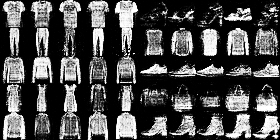}};
    \draw[white] (.0,-2.5) -- (.0,2.5);
    \end{tikzpicture}}
    \caption{Fashion MNIST}\label{fig:sampling_fashionmnist}
    \end{subfigure}
    \caption{LU-Net: Randomly generated samples of (a) MNIST and (b) Fashion MNIST. These examples are generated by sampling random numbers from a multivariate normal distribution and passing them through the inverse of LU-Net. Moreover, these images are ordered class-wise and in decreasing likelihood from left to right as estimated by LU-Net.}
    \label{fig:sampling_mnist_fashionmnist}
\end{figure}

\begin{figure}
    \centering
    \begin{subfigure}{.24\textwidth}
    \resizebox{\linewidth}{!}{
       \begin{tikzpicture}
        \node[inner sep=0pt] at (0,0) {\includegraphics{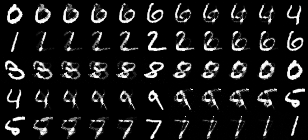}};
        \draw[red, line width=2pt] (-5.4,-2.5) rectangle (-4.5,2.5);
        \draw[red, line width=2pt] (-.5,-2.5) rectangle (.5,2.5);
        \draw[red, line width=2pt] (4.5,-2.5) rectangle (5.4,2.5);
    \end{tikzpicture}}
    \caption{MNIST}\label{fig:interpolation_mnist}
    \end{subfigure}%
    \begin{subfigure}{.24\textwidth}
    \resizebox{\linewidth}{!}{
   \begin{tikzpicture}
        \node[inner sep=0pt] at (0,0) {\includegraphics{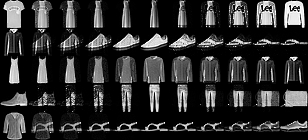}};
        \draw[red, line width=2pt] (-5.4,-2.5) rectangle (-4.5,2.5);
        \draw[red, line width=2pt] (-.5,-2.5) rectangle (.5,2.5);
        \draw[red, line width=2pt] (4.5,-2.5) rectangle (5.4,2.5);
    \end{tikzpicture}}
    \caption{Fashion MNIST}\label{fig:interpolation_fashionmnist}
    \end{subfigure}
    \caption{LU-Net: Samples of (a) MNIST and (b) Fashion MNIST generated by interpolating in latent space of LU-Net. The examples in the red boxes are reconstructions from latent representations of original test data. Note that the examples are ordered randomly.}
    \label{fig:interpolation_mnist_fashionmnist}
\end{figure}
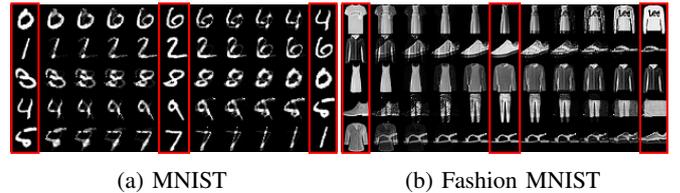

In \Cref{tab:nll-mnist} we report the negative log likelihood computed by LU-Net on the MNIST and Fashion-MNIST test datasets. In \Cref{fig:density_mnist_fashionmnist} we provide qualitative examples of LU-Net as density estimator and in \Cref{fig:sampling_mnist_fashionmnist} as well as \Cref{fig:interpolation_mnist_fashionmnist} as generator of new samples.

In terms of numerical results, we achieve adequate density estimation scores with an average NLL of $2.75 \pm 0.29$ bits/pixel and $3.74 \pm 2.46$ bits/pixel over all classes on MNIST and Fashion-MNIST, respectively. In general, the results can be considered as robust for each class of both datasets. The only major deviations are related to the classes T-Shirt and Trousers of Fashion-MNIST, which can be explained by the large variety of patterns in different examples and hence the increased difficulty in generative modeling, \cf also \Cref{tab:nll-mnist} {\color{red}right} and \Cref{fig:density-fashionmnist}. Furthermore, we notice that LU-Net is capable of assigning meaningful likelihoods, \ie images that are more characteristic for the associated class are assigned higher likelihoods, see again \Cref{fig:density_mnist_fashionmnist} in particular.

With regard to LU-Net as generator, we obtain reasonable quality of the sampled images. The random samples can clearly be recognized as subset of MNIST or Fashion-MNIST and moreover, they can also be assigned easily to the corresponding classes. However, we notice that many generated examples contain noise, most visible for class 7 in MNIST or class sandal and bag in Fasion-MNIST, \cf \Cref{fig:sampling_mnist} and \Cref{fig:sampling_fashionmnist}, respectively. 
This shortcoming is not surprising since the fully connected layers of LU-Net capture less spatial correlations as other filter based architectures commonly used on image data.

Another noteworthy observation refers to the learned latent space of LU-Net, \ie the space after applying the normalizing sequence. Given its invertibility property, each latent variable represents exactly one image, which allows for traveling though the latent space and thus also interpretation of it. By interpolating between two latent representations, we generally observe a smooth transition between the two corresponding images when transforming back to original input space, see \Cref{fig:interpolation_mnist} for MNIST and \Cref{fig:interpolation_fashionmnist} for Fashion-MNIST. This also enables the visual inspection of relevant features or parts of the content associated with certain images or classes.

To conclude, we have seen that LU-Net even with a shallow architecture can be applied as probabilistic generative model to images. The numerical results on these higher dimensional data indicate that the bijectivity constraint is not a significant limitation regarding the expressive power. Although not specifically designed to model image data, LU-Net is still capable of generating sufficiently clear images, which highlights its general purpose property as generative model.

\begin{table}
\centering
\begin{tabular}{|c|c|c|c|}
    \hline
    \multicolumn{2}{|c|}{MNIST Test NLL} & \multicolumn{2}{|c|}{Fashion-MNIST Test NLL} \\
    \hline\hline
    Class & Bits / Pixel $\downarrow$ & Class & Bits / Pixel $\downarrow$ \\
    \hline
    Number 0 & 5.2647 $\pm$ 0.0189  & T-Shirt    & 6.0803 $\pm$ 0.0321 \\
    Number 1 & 5.0888 $\pm$ 0.1320  & Trousers   & 5.5130 $\pm$ 0.0778 \\
    Number 2 & 6.0798 $\pm$ 0.0249  & Pullover   & 6.1770 $\pm$ 0.0043 \\
    Number 3 & 5.0174 $\pm$ 1.0864  & Dress      & 6.0273 $\pm$ 0.1005 \\
    Number 4 & 5.0952 $\pm$ 0.0278  & Coat       & 6.1355 $\pm$ 0.0515 \\
    Number 5 & 5.4285 $\pm$ 0.0742  & Sandal     & 5.9324 $\pm$ 0.0449 \\
    Number 6 & 5.4642 $\pm$ 0.0393  & Shirt      & 6.2517 $\pm$ 0.0437 \\
    Number 7 & 5.7739 $\pm$ 0.0040  & Sneaker    & 5.6603 $\pm$ 0.0007 \\
    Number 8 & 5.4086 $\pm$ 0.0555  & Bag        & 6.3035 $\pm$ 0.0478 \\
    Number 9 & 5.0607 $\pm$ 0.0012  & Ankle Boot & 5.8741 $\pm$ 0.0314 \\
    \hline\hline
    Average & 5.3682 $\pm$ 0.1464 & Average & 5.9956 $\pm$ 0.0386 \\
    \hline
\end{tabular}
\caption{Class-wise negative log likelihood in bits per pixel when applying RealNVP to MNIST and Fashion-MNIST test dataset. The results are averaged over 30 runs.}
\label{tab:nllnvp}
\end{table}

\subsection{Comparison with RealNVP}

\begin{table}
\centering
\setlength{\tabcolsep}{4pt}
\begin{tabular}{|c|c|c|c|c|}
    \hline
            & num weight    & GPU memory    & num epochs    & test NLL    \\
    model   & parameters    & usage in MiB  & training      & in bits/pixel      \\
    \hline
    \hline
    LU-Net  & 4.92M & 1,127 & 40    & 3.2424 \\
    RealNVP & 5.39M & 3,725 & 100   & 5.6819 \\
    \hline
    \multicolumn{5}{c}{} \\
    \hline
            & train epoch   & optimization  & density per   & sampling per  \\
    model   & in sec        & step in ms   & image in ms   & image in ms   \\
    \hline
    \hline
    LU-Net  & 7.32  & 1.2 & 37.10  & 45.15  \\
    RealNVP & 99.88 & 56.0 & 259.15 & 1.03   \\
    \hline
\end{tabular}
\caption{Model size and run time comparisons between LU-Net and RealNVP. Note that the time results are averages over 100 runs on MNIST and Fashion-MNIST. Moreover, all these tests were conducted on the same machine using an NVIDIA Quadro RTX 8000 GPU and a batch size of 128.}
\label{tab:comparison-lunet-realnvp}
\end{table}

In these final experiments we want to compare LU-Net with the popular and widely used normalizing flow architecture RealNVP \cite{dinh2017density}. To make the models better comparable, we design them to be of similar size in terms of model parameters. In more detail, we employ a RealNVP normalizing flow with 9 affine coupling layers with checkerboard mask mixing. For the coupling layers two small ResNets \cite{he2016deep} are used to compute the scale and translation parameter, respectively. Here, every ResNet backbone consists of two residual blocks, each applying two $3 \times 3$ convolutions with 64 kernels and ReLU activations. In total this amounts to a normalizing flow with 5,388,264 weight parameters compared to 4,920,384 weight parameters with LU-Net. Finally, we conducted the same experiments as presented in \Cref{sec:exp_mnist} for RealNVP. 

In \Cref{tab:nllnvp} we report the negative log likelihood scores of our implemented RealNVP on the test splits of MNIST as well as Fashion-MNIST. In comparison to LU-Net, we observe more robust results but overall significantly worse performance in density estimation for each class of both datasets with RealNVP. During the experiments with RealNVP, we realized that the model needs to treated carefully as slight modifications in the hyper parameters could quickly lead to unstable training. We ended up training the flows for 100 epoch using a small learning rate of 1e-4 that decays by 0.2 every 10 epochs.

Besides the worse performance on learning the data distribution of MNIST and Fashion-MNIST, RealNVP further is computationally more expensive than LU-Net, which can be seen in \Cref{tab:comparison-lunet-realnvp} showing a comparison of the computational budget. RealNVP not only requires more GPU memory than LU-Net but also considerably more time to train. The latter point can be explained by backpropagation not working as efficiently due to the deep neural networks employed in coupling layers, which adversely affects propagating the errors from layer to layer. Moreover, RealNVP is notably slower at density evaluation, with LU-Net being 7 times faster. With regard to sampling, RealNVP is however significantly faster by nearly 50 times. Here, we want to note that at run time linear equation systems are solved in inverse LU-Net instead of inverting the weight matrices, \cf \Cref{sec:inverse-lu-net} and \Cref{sec:reverse-function}, respectively. Although the inversion could be performed offline saving a considerable amount of operations, the computation of big inverse matrices is often numerically unstable, which is highly undesirable in particular in the context of invertible neural networks and therefore omitted. 

\begin{figure}
\centering
\begin{subfigure}{0.49\linewidth}
    \includegraphics[width=\textwidth]{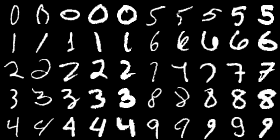}
    \caption{MNIST}
    \label{fig:OriginalFlowMNIST}
\end{subfigure}
\begin{subfigure}{0.49\linewidth}
    \includegraphics[width=\textwidth]{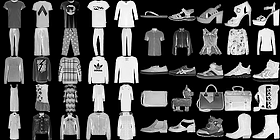}
    \caption{Fashion MNIST}
    \label{fig:OriginalFlowFashion}
\end{subfigure}
\caption{Unseen original test samples of (a) MNIST and (b) Fashion MNIST. These images are ordered class-wise and in decreasing likelihood from left to right as estimated by RealNVP. Note that RealNVP incorrectly assigns high likelihoods to some unclear examples, and vice versa.} 
\label{fig:density_realnvp}
\end{figure}

\begin{figure}
\centering
\begin{subfigure}{0.49\linewidth}
    \includegraphics[width=\textwidth]{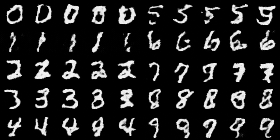}
    \caption{MNIST}
    \label{fig:GenerationFlowMNIST}
\end{subfigure}
\begin{subfigure}{0.49\linewidth}
    \includegraphics[width=\textwidth]{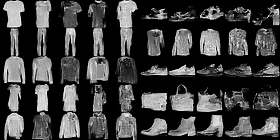}
    \caption{Fashion MNIST}
    \label{fig:GenerationFlowFashion}
\end{subfigure}
\caption{RealNVP: Randomly generated samples of (a) MNIST and (b) Fashion MNIST. These images are ordered class-wise and in decreasing estimated likelihood from left to right.} 
\label{fig:figFlowSamplings}
\end{figure}

Lastly, we present qualitative examples of RealNVP as density estimator in \Cref{fig:density_realnvp} and as generator in \Cref{fig:figFlowSamplings} for MNIST and Fashion-MNIST. At first glance, it is directly notable that the generated images are less noisy in comparison to the images generated by LU-Net. This might be an consequence of the extensive use of convolution operations in RealNVP that helps the model to better capture local correlations of features in images. However, the shapes of the digits and clothing articles in the images generated by our implemented slim RealNVP are still rather unnatural, which can be solved by deeper RealNVP models \cite{dinh2017density}.

\section{Conclusion and Outlook} \label{sec:conclusion}
We introduced LU-Net, which is a simple architecture for an invertible neural network based on the LU-factorization of weight matrices and invertible as well as two times differentiable activation functions. LU-Net provides an explicit likelihood evaluation and reasonable sampling quality. The execution of both tasks is computationally cheap and fast, which we tested in several experiments on academic data sets.

In the future, we intend to investigate more closely the effects of the choice of the activation function. 
 LU-net would become even simpler if  leaky ReLu activation functions could be used, which can be achieved by  training adversarily. Also, we intend to revisit the universal approximation properties for LU-Nets using artificial widening via zero padding.

\section*{Acknowledgment}
 This work has been funded by the German Federal Ministry for Economic Affairs and Climate Action (BMWK) via the research consortium AI Delta Learning (grant no.\ 19A19013Q) and the Ministry of Culture and Science of the German state of North Rhine-Westphalia as part of the KI-Starter research funding program (grant no. 005-2204-0023). Moreover, this work has been supported by the German Federal Ministry of Education and Research (grant no.\ 01IS22069).

\bibliography{references} 
\bibliographystyle{ieeetr}

\clearpage
\onecolumn
\appendix

\subsection{Condition numbers} \label{sec:condition}
	To make the inversion of the LU network possible, it is necessary, that the L and U weight matrices are well conditioned. The condition number $\mathsf{\kappa}$ of a matrix $\mathsf{A\in\mathbb{R}^{D\times D}}$ is defined as
	\begin{equation*}
		\mathsf{\kappa(A)=||A||_2\cdot||A^{-1}||_2}
	\end{equation*}	
    and it is a measure for the numeric stability of a linear equation system $\mathsf{A\cdot x=b}$ for any vector $\mathsf{b\in\mathbb{R}^D}$. That is why we use the condition numbers of the weight matrices as a measure for the success of the inversion of the LU network.

     The results for the MNIST and FashionMNIST datasets are  reported in Tables \ref{tab:figMNISTCondNumbers} and \ref{tab:figFashionMNISTCondNumbers}.
	
\begin{table}[h]
    \centering
    \begin{tabular}{|c|c|c|c|c|c|c|c|c|}
    \hline
    \multicolumn{9}{|c|}{Condition numbers} \\
    \hline
    \text{ } & $\mathsf{\kappa(U^{(1)})}$ &$\mathsf{\kappa(L^{(1)})}$&$\mathsf{\kappa(U^{(2)})}$ & $\mathsf{\kappa(L^{(2)})}$&$\mathsf{\kappa(U^{(3)})}$&$\mathsf{\kappa(L^{(3)})}$&$\mathsf{\kappa(U^{(4)})}$&$\mathsf{\kappa(L^{(4)})}$\\
    \hline
    Untrained weights & $6.94e+18$ & $3.18$ & $2.55e+18$ & $3.19$ & $8.44e+17$ & $3.18$ & $1.89e+18$ & $3.16$\\
    \hline
    Number 0 & $2.13$ & $690.39$ & $3.04$ & $138.01$ & $2.66$ & $50.59$ & $2.40$ & $271.66$\\
    Number 1 & $2.39$ & $456.42$ & $3.01$ & $195.82$ & $2.70$ & $51.40$ & $2.53$ & $216.44$\\
    Number 2 & $2.14$ & $321.54$ & $3.51$ & $199.98$ & $2.69$ & $54.75$ & $2.43$ & $190.21$\\
    Number 3 & $2.13$ & $318.00$ & $3.35$ & $184.68$ & $2.77$ & $61.91$ & $2.61$ & $161.94$\\
    Number 4 & $2.16$ & $618.06$ & $2.86$ & $298.37$ & $2.59$ & $41.72$ & $2.39$ & $255.30$\\
    Number 5 & $2.07$ & $177.44$ & $2.95$ & $131.62$ & $2.51$ & $40.94$ & $2.43$ & $248.46$\\
    Number 6 & $2.11$ & $300.60$ & $3.10$ & $179.20$ & $2.49$ & $43.89$ & $2.39$ & $236.14$\\
    Number 7 & $2.45$ & $431.88$ & $3.50$ & $312.08$ & $2.87$ & $80.46$ & $2.52$ & $188.26$\\
    Number 8 & $2.16$ & $433.54$ & $2.94$ & $340.09$ & $2.64$ & $43.51$ & $2.51$ & $184.29$\\
    Number 9 & $2.19$ & $743.07$ & $3.15$ & $149.99$ & $2.53$ & $50.74$ & $2.40$ & $205.62$\\
    \hline
    \end{tabular}
    \caption{MNIST: Condition numbers of weight matrices before and after the training} 
    \label{tab:figMNISTCondNumbers}
\end{table}

\begin{table}[h]
    \centering
    \begin{tabular}{|c|c|c|c|c|c|c|c|c|}
    \hline
    \multicolumn{9}{|c|}{Condition numbers} \\
    \hline
    \text{ } & $\mathsf{\kappa(U^{(1)})}$ &$\mathsf{\kappa(L^{(1)})}$&$\mathsf{\kappa(U^{(2)})}$ & $\mathsf{\kappa(L^{(2)})}$&$\mathsf{\kappa(U^{(3)})}$&$\mathsf{\kappa(L^{(3)})}$&$\mathsf{\kappa(U^{(4)})}$&$\mathsf{\kappa(L^{(4)})}$\\
    \hline
    Untrained weights & $6.94e+18$ & $3.18$ & $2.55e+18$ & $3.19$ & $8.44e+17$ & $3.18$ & $1.89e+18$ & $3.16$\\
    \hline
    T-Shirt & $2.38$ & $177.70$ & $3.26$ & $146.45$ & $2.85$ & $52.19$ & $2.56$ & $214.36$\\
    Trouser & $2.51$ & $727.66$ & $2.96$ & $224.79$ & $2.63$ & $48.70$ & $2.46$ & $191.40$\\
    Pullover & $2.26$ & $202.77$ & $2.94$ & $105.24$ & $2.75$ & $49.84$ & $2.54$ & $217.66$\\
    Dress & $2.34$ & $621.33$ & $3.14$ & $165.46$ & $2.80$ & $56.55$ & $2.57$ & $263.57$\\
    Coat & $2.25$ & $244.03$ & $3.06$ & $149.92$ & $2.73$ & $57.00$ & $2.57$ & $275.77$\\
    Sandal & $2.23$ & $521.50$ & $3.12$ & $208.06$ & $3.01$ & $52.25$ & $2.55$ & $150.41$\\
    Shirt & $2.41$ & $168.12$ & $2.94$ & $104.54$ & $2.93$ & $52.73$ & $2.55$ & $235.12$\\
    Sneaker & $1.68$ & $9.26$ & $1.42$ & $3.88$ & $1.51$ & $4.07$ & $1.61$ & $59.43$\\
    Bag & $2.71$ & $238.70$ & $3.26$ & $109.81$ & $2.78$ & $58.26$ & $2.64$ & $210.89$\\
    Ankle Boot & $1.78$ & $9.53$ & $1.46$ & $4.06$ & $1.51$ & $4.32$ & $1.66$ & $58.18$\\
    \hline
    \end{tabular}
    \caption{Fashion-MNIST: Condition numbers of weight matrices before and after the training}
    \label{tab:figFashionMNISTCondNumbers}
\end{table}

\subsection{Test for normality} \label{sec:normality}

\begin{figure}[h]
\centering
\begin{subfigure}{1.0\linewidth}
    \includegraphics[width=\textwidth]{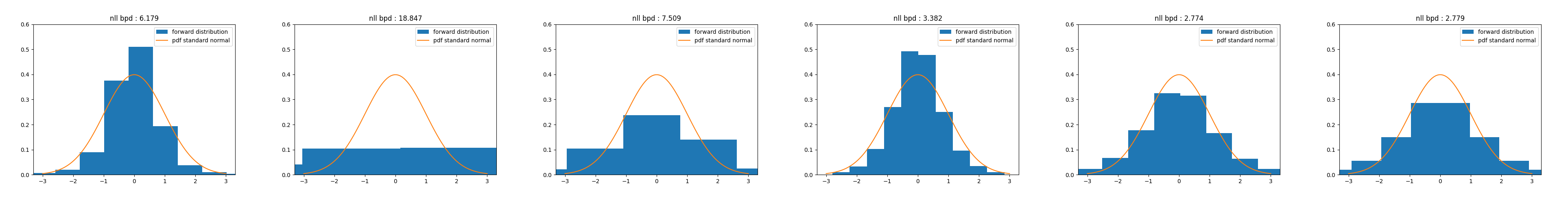}
    \caption{MNIST: Number 3}
\end{subfigure}
\begin{subfigure}{1.0\linewidth}
    \includegraphics[width=\textwidth]{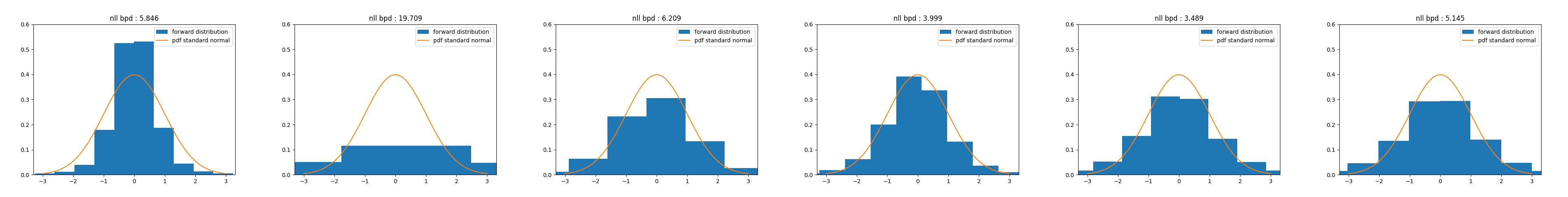}
    \caption{Fashion MNIST: Bag}
\end{subfigure}
\caption{Distribution of 1200 test data while training after epoch 1, 3, 6, 8, 15 and 30 (from left to right) with the negative log likelihood in bits per pixel as title of each plot.}
\label{fig:figDistribution}
\end{figure}

INNs are trained  to map the data distribution onto a multivariate standard normal distribution in the normalizing direction. We therefore inspect the success of the normalizing direction by visual comparison while (\Cref{fig:figDistribution}) and after training. 

It is well known that $\mathsf{Z\sim N(0,\mathds{1})}$ in $\mathsf{D}$ dimensions, if and only if $\mathsf{c^\top Z\sim N(0,1)}$ for all $\mathsf{c\in \mathbb{R}^D}$ with $\mathsf{\|c\|=1}$. 

We thus test normality by sampling random directions $\mathsf{c}$, normalize $\mathsf{ c\mapsto \nicefrac{c}{\|c\|}}$ and collect values $\mathsf{z^{(n)}=c^\top f(x^{(n)})}$, $\mathsf{n=1,\ldots,N}$ where $\mathsf{f}$ is the trained LU-Net trained evaluated on the test data set $\mathsf{\{x_n\}_{n=1}^N}$. We thereafter display a histogram of the $\mathsf{z^{(n)}}$-values and compare it to the density of the standard normal distribution. 

\begin{figure}[h]
\centering
\begin{subfigure}{1.0\linewidth}
    \includegraphics[width=\textwidth]{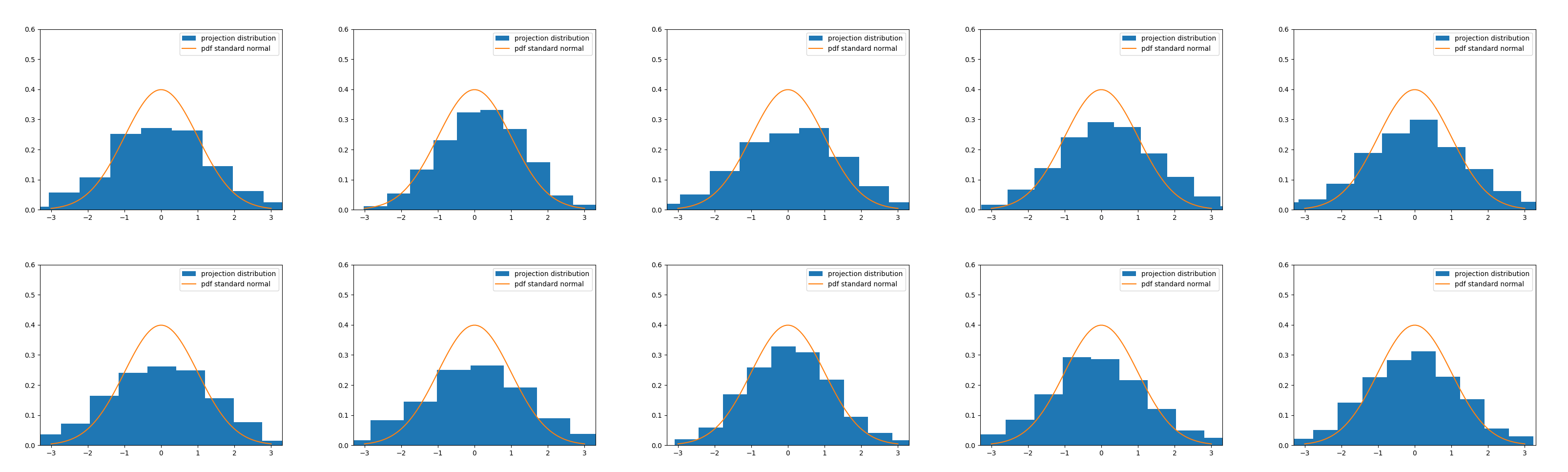}
\end{subfigure}
\caption{MNIST: Randomly chosen projection for number 0, 1, 2, 3 and 4 in the first row (from left to right) and number 5, 6, 7, 8 and 9 in the second row (from left to right). In total 2000 test data were used for each projection.}
\label{fig:figProjectionMNIST}
\end{figure}

\begin{figure}[h]
\centering
\begin{subfigure}{1.0\linewidth}
    \includegraphics[width=\textwidth]{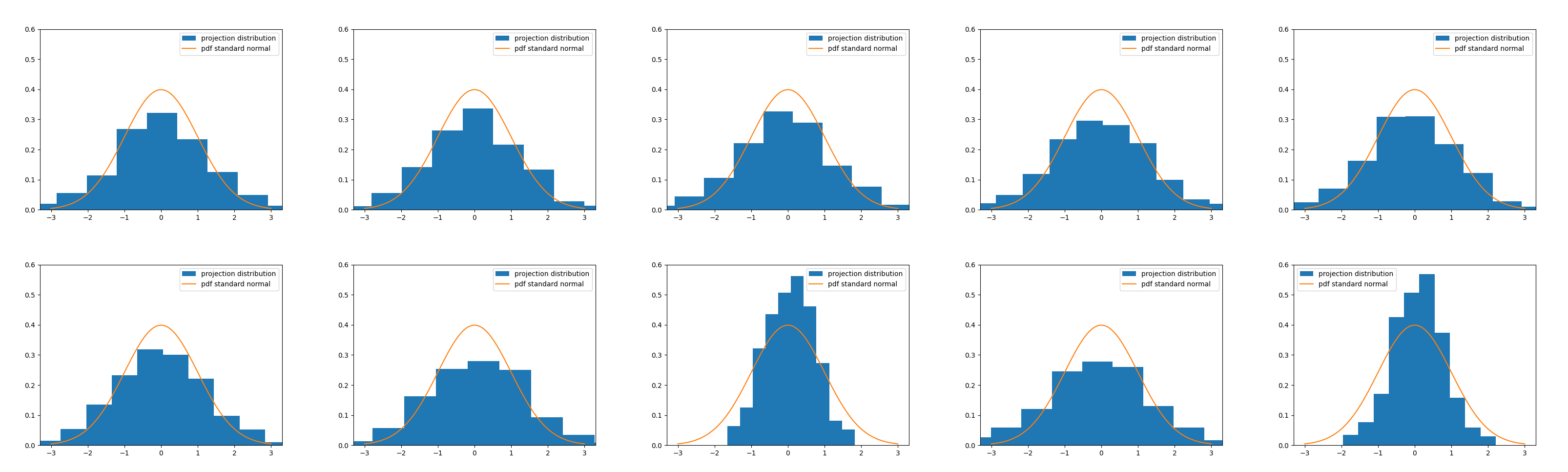}
\end{subfigure}
\caption{FashionMNIST: Randomly chosen projection for T-Shirt, Trouser, Pullover, Dress and Coat in the first row (from left to right) and Sandal, Skirt, Sneaker, Bag and Ankle Boot in the second row (from left to right). In total 2000 test data were used for each projection.}
\label{fig:figProjectionFashionMNIST}
\end{figure}

Figures \ref{fig:figProjectionMNIST} and \ref{fig:figProjectionFashionMNIST} display the result for one independently sampled direction $\mathsf{c}$ of projection for each of the classes of MNIST and Fashion-MNIST, respectively.   

\clearpage
\subsection{Inverted forward function} \label{sec:reverse-function}
	For the numerical stability, it is not beneficial, when we invert the weight matrices of the L and U layer in the forward function of an inverted LU layer. Firstly, we solve the linear equation system
	\begin{equation*}
		\mathsf{Ly=\left(\Phi^{-1}(z)-b\right)^\intercal}
	\end{equation*}
	in the inverted L layer. Secondly in the inverted U layer, we calculate the solution of following linear system
	\begin{equation*}
		\mathsf{Rx=y},
	\end{equation*}
which is computationally more expensive than $\mathsf{x=R^{-1}y}$ but numerically more stable.

\end{document}